\documentclass[runningheads]{llncs}
\usepackage[T1]{fontenc}
\usepackage{graphicx}

\usepackage{amsmath}
\usepackage{amssymb}
\usepackage{xspace}
\usepackage{xcolor}
\usepackage{dsfont}
\usepackage{rotating}
\usepackage{makecell}
\usepackage{multirow}
\usepackage{tabularx}
\usepackage{subcaption}
\usepackage{placeins,booktabs}
\usepackage[export]{adjustbox}% http://ctan.org/pkg/adjustbox
\usepackage[colorinlistoftodos,prependcaption]{todonotes}
\usepackage{xargs}   % Use more than one optional parameter in a new commands

\usepackage{pifont}          % for check- and cross-marks
\newcommand{\cmark}{\ding{51}} % ✓
\newcommand{\xmark}{\ding{55}} % ✗

\usepackage{algorithm}
\usepackage{algpseudocode,soul}

\newcommandx{\go}[2][1=]{\todo[inline,linecolor=orange,backgroundcolor=orange!25,bordercolor=orange,#1]{\textbf{Gabriela:}#2}}

\newcommandx{\ns}[2][1=]{\todo[inline,linecolor=blue,backgroundcolor=blue!25,bordercolor=blue,#1]{\textbf{Niki:}#2}}

\newcommandx{\hy}[2][1=]{\todo[inline,linecolor=green,backgroundcolor=green!25,bordercolor=green,#1]{\textbf{Haoran:}#2}}

\newcommandx{\ak}[2][1=]{\todo[inline,linecolor=red,backgroundcolor=red!25,bordercolor=red,#1]{\textbf{Anna:}#2}}

\begin{document}
\title{Behaviour Space Analysis\\ of LLM-driven Meta-heuristic Discovery}
\author{
%Anonymous authors
Niki van Stein\inst{1}\orcidID{0000-0002-0013-7969} \and 
Haoran Yin\inst{1} \and 
Anna V. Kononova\inst{1}\orcidID{0000-0002-4138-7024} 
Thomas B\"ack\inst{1}\orcidID{0000-0001-6768-1478} \and 
Gabriela Ochoa\inst{2}\orcidID{0000-0001-7649-5669}
}
\authorrunning{N. van Stein et al.}
\institute{Leiden University, Leiden, The Netherlands, \email{n.van.stein@liacs.leidenuniv.nl} \and University of Stirling, Scotland, UK
}
\maketitle              % typeset the header of the contribution
\begin{abstract}
%\hl{deadline 03.07} 
We investigate the behaviour space of meta-heuristic optimisation algorithms automatically generated by
Large Language Model driven algorithm discovery methods. Using the Large Language Evolutionary Algorithm (LLaMEA) framework with a GPT o4-mini LLM, we iteratively evolve
black-box optimisation heuristics, evaluated on 10 functions from the BBOB benchmark suite. Six LLaMEA variants,
featuring different mutation prompt strategies, are compared and analysed. We log dynamic behavioural metrics including exploration, exploitation, convergence and stagnation measures, for each run, and analyse
these via visual projections and network-based representations. Our analysis combines behaviour-based
 projections, Code Evolution Graphs built from static code features, performance
convergence curves, and behaviour-based Search Trajectory Networks. The results reveal clear differences in search dynamics and algorithm structures across LLaMEA configurations. Notably, the variant that
employs both a code simplification prompt and a random perturbation prompt in a 1+1 elitist evolution
strategy, achieved the best performance, with the highest Area Over the Convergence Curve.
Behaviour-space visualisations show that higher-performing algorithms exhibit more intensive exploitation behaviour and faster convergence with less stagnation. Our findings demonstrate how behaviour-space analysis can explain why certain LLM-designed heuristics outperform others and how LLM-driven algorithm discovery navigates the open-ended and complex search space of algorithms. These findings provide insights to guide the future design of adaptive LLM-driven algorithm generators.
\end{abstract}
\section{Introduction}

The automated design of optimisation algorithms, often termed \emph{Automated Algorithm Design} (AAD) \cite{stutzle2018automated}, has
been a longstanding challenge in evolutionary computation and the field of AI. Early approaches like \textit{Programming by optimisation} (PbO) \cite{hoos2012programming} advocated maintaining rich sets of algorithm components to be tuned
by meta-optimisation methods. Subsequent research introduced dynamic algorithm configuration and modular
algorithm frameworks such as modular CMA-ES \cite{modcma} and modular DE \cite{modDE23}, evolving algorithm structures for better performance. With the advent of
powerful \textit{Large Language Models} (LLMs), a new paradigm has emerged: using LLMs to generate and evolve functions or entire algorithms in natural language and code \cite{chauhan2025evolutionary,liu2024systematic}. Recent work has shown LLMs can
automatically produce novel meta-heuristics that rival human-designed ones \cite{pluhacek2023leveraging,vanstein2024llamealargelanguagemodel}. For example, the
LLaMEA framework by van Stein et al. \cite{vanstein2024llamealargelanguagemodel} uses an LLM in-the-loop to iteratively generate and improve
black-box optimizers, and other studies have explored evolving heuristics with LLMs in various contexts \cite{fei2024eoh,FunSearch2024,ye2024reevo,van2024loop}.

However, a key challenge remains: \textbf{understanding the behaviour} of these LLM-generated algorithms. While
performance comparisons (e.g., final accuracy or convergence speed) tell us which method works best, they
do not explain why. Traditional algorithm analysis tools, such as fitness landscape analysis and
exploratory landscape analysis \cite{mersmann2011exploratory}  for problem instances, are not directly applicable when the ``individuals'' being evolved are \textit{algorithms themselves}. Currently, we have limited insight into how an LLM-driven algorithm
search navigates the space of possible programs, how algorithm structure and strategy evolve over time,
and what behavioural characteristics distinguish successful runs from failures. Gaining such insights
is crucial for \textit{explainability} and for improving the automated design process \cite{vanstein2024explainable}.

In this paper, we address this gap by conducting a comprehensive behaviour space analysis of LLM-generated
meta-heuristics. We focus on the LLaMEA framework using an OpenAI GPT-based model (\textit{o4-mini-2025-04-16}) to evolve algorithms for continuous optimisation. We benchmark six different LLaMEA configurations on a set of $10$ noiseless BBOB functions \cite{bbob_hansen2009_noiseless}, and record entire traces from each run using IOH experimenter \cite{IOHexperimenter}. By computing behavioural metrics from these traces and leveraging advanced visualisation techniques, we reveal patterns in exploration, exploitation, convergence, and stagnation exhibited by the generated algorithms.

Our contributions are as follows: 
\begin{itemize}
    \item We define a set of quantitative behaviour metrics for LLM-generated optimizers, capturing aspects such as search space coverage, intensification near optima, convergence speed, and stagnation periods.
    \item We use Parallel coordinate plots, code evolution graphs and behaviour-based search trajectory networks for visual comparison of algorithm ``footprints'' in behaviour space.
    \item We compare $6$ different variants of mutation and evolutionary search strategies and identify which configuration yields the best performance (via AOCC metric) and use our behaviour analysis to explain its success. In particular, we find that a 1+1 elitist strategy combining simplification and random perturbation prompts consistently outperforms others, likely due to a balanced exploration-exploitation and sustained improvement without disruption.
\end{itemize}

%The rest of the paper is organized as follows. ...
%TODO

\section{Related Work}

\subsection{Automated Algorithm Design}

The concept of automatically designing or configuring algorithms has evolved over the past decade. The \textit{Programming by optimisation} paradigm \cite{hoos2012programming} encouraged specifying algorithm components as parameters to be tuned rather than fixed, enabling automated search
in algorithm space. A related approach, \textit{Genetic Improvement} \cite{langdonH2015}, attempts to automatically improve the behaviour of a software system using evolutionary algorithms. Starting from human-written code, genetic improvement tries to evolve it so that it is better with respect to
given criteria, typically non-functional properties, such as execution time and power consumption, though others are possible.
%\go{Added a sentence on Genetic Improvement above, but not sure if it is relevant? as it starts from humman written code.}

This idea was expanded to numerical optimisation through modular algorithm frameworks that assemble heuristics from predefined components. Notably, van Rijn et al. \cite{vanRijn2016} demonstrated evolving the structure of CMA-ES variants via evolutionary strategies, effectively treating algorithm blueprint as a genotype. These earlier approaches required a human-defined modular design space, whereas the rise of LLMs enables generating algorithms free-form, without a fixed module library. The works of Liu et al. \cite{fei2024eoh} and Zhang et al. \cite{zhang2024understanding} exemplify this new direction: by leveraging an LLM (GPT-style) to propose pseudocode or code for heuristics, they achieved competitive optimisation results and underscored the importance of evolutionary search in guiding LLMs. The LLaMEA framework and related efforts have further formalized LLM-driven algorithm evolution as a novel class of evolutionary computation technique.

\subsection{Analysing Meta-heuristic Behaviour}
Understanding why an optimisation algorithm succeeds or fails on a problem has long been of interest. Traditionally, researchers have analysed algorithm trajectories through
frameworks like fitness landscape analysis and performance footprints. 

%\go{Added a paragraph about STNs below}
\paragraph{Search Trajectory Networks (STNs)} \cite{OCHOA2021107492} are a data-driven, graph-based tool to visualize,  quantify and contrast the dynamics of iterative optimisation algorithms. In an STN, nodes are \textit{locations} of \textit{representative solutions} in the search space, and edges connect successive locations in the search trajectory. Edges are weighted with the sampling frequency of transitions between pairs of nodes during the STN construction process, which is based on logs from a number of runs of the studied algorithms on the considered problems. A representative solution is a solution to the optimisation problem at a given time step that represents the status of the search algorithm, for example, the best in the population, although all solutions could be logged for small populations. In the initial STN model \cite{OCHOA2021107492}, representative solutions belong to the genotype search space. For complex search spaces (i.e neural networks, graphs, programs), solutions can be logged in the phenotype or behaviour spaces \cite{hu2023phenotype,NadizarRMO2025,sarti2022neuroevolution}, which are generally much smaller. Since any search space for realistic optimisation problems is very large, a process of partitioning the space into locations is required to have manageable models. A partition strategy groups subset of related solutions into locations.  For example, if solutions are vectors of real numbers, locations can be defined as hypercubes with a prefixed lower numerical precision~\cite{OCHOA2021107492}. Each solution is an element of one and only one location, and each location is assigned a representative objective/fitness value.

%\ak{Refine / rewrite the Attractor networks part}
Attractor Networks (ANs) \cite{thomson2025stalling} were later developed by Thomson et al. to highlight regions where an optimizer stagnates. These represent the ``stalling'' behaviour by grouping contiguous non-improving iterations
into nodes (attractors). Another related concept is Local Optima Networks (LONs) \cite{ochoa2014local}, which map the connectivity of local optima in combinatorial landscapes. LONs and ANs provide coarse views of search dynamics but have mainly been applied to fixed algorithms. Here, we extend the idea to algorithm-evolution processes, where the ``search'' happens over algorithm designs.

\subsection{Explainable Benchmarking}
%\ak{Read / improve this section please}
While performance traces reveal how an algorithm searches, static code features reveal what the algorithm is. Recent work has explored using code analysis for understanding algorithm performance. Pulatov et al. \cite{pulatov2022opening} showed that static metrics like cyclomatic
complexity and Abstract Syntax Tree (AST) structure can improve algorithm selection by characterising algorithm behaviour beyond external performance alone. In the context of AAD, static analysis can help quantify how algorithms generated by LLMs differ structurally. For example, van Stein et al. \cite{van2025code} introduced \emph{Code Evolution Graphs} (CEGs) as a method to trace the lineage of evolving code and its properties. CEGs integrate graph metrics (from ASTs) and code complexity measures to visualise algorithm evolution over generations. This approach is inspired by earlier software visualisation techniques like Code Flow graphs \cite{telea2008code}, which depict how code changes across versions. To our knowledge, no prior work has systematically combined dynamic behaviour analysis (e.g. STNs, attractors) with static code analysis in the AAD domain. By doing so, our study provides a \textit{holistic view} of LLM-generated meta-heuristics, linking how they search to what they consist of and how they perform.

Finally, our work is related to efforts in explainable AI and benchmarking for
optimisation heuristics. The BLADE benchmark suite \cite{van2025blade} was recently proposed to evaluate LLM-driven AAD methods in a standardized way, including logging of all runs for transparency. In particular, metrics like the \textit{Area Over the Convergence Curve} (AOCC) \cite{hansen2022anytime} have been promoted as a more informative performance measure than final results alone, as they capture anytime performance. We adopt AOCC in our comparisons to evaluate not just whether an algorithm eventually finds a good optimum, but how
quickly and consistently it does so over the evaluation budget. Our analysis framework can be seen as contributing to \textit{explainable benchmarking}: by augmenting performance metrics with behavioural and structural analysis, we can better interpret the outcomes of an algorithm design experiment.

\paragraph{Putting it together.}
To our knowledge no prior work combines (i) dynamic search‐trace
analysis (STNs/ANs), (ii) static code metrics (CEGs), and (iii)
standardised anytime performance measures such as AOCC
within a single benchmark.
Our methodology closes that triangle, providing the first
holistic view of LLM-driven automated algorithm design.

\section{Methodology}

%\hy{Read / improve this section where needed}

\subsection{LLaMEA Framework and Configurations}

We build our study on \textbf{LLaMEA} \cite{vanstein2024llamealargelanguagemodel} – a framework where an LLM acts as a generative engine to produce and iteratively refine algorithms. In LLaMEA, the individuals in the evolutionary process are complete algorithms (expressed as code).
Each algorithm individual is evaluated by running it on a user-defined evaluation function, in this case a set of optimisation problems, to obtain a performance score (e.g., best function value found, in our case the any-time performance metric AOCC averaged over multiple instances and runs). \textit{Selection} then chooses algorithms for the next iteration, and variation is implemented by prompting the LLM to generate modified algorithms (analogous to \textit{mutation/crossover} in genetic algorithms).
In our experiments, we use an OpenAI GPT-derived model named \textit{o4-mini-2025-04-16} as the generative LLM. This is a relatively compact LLM specialized for code generation and LLM-reasoning \cite{plaat2024reasoning}, allowing multiple calls within our computational budget.
We integrate LLaMEA with the BLADE benchmarking framework \cite{van2025blade} to ensure rigorous evaluation and logging. The target problems are $10$ noiseless BBOB functions \cite{bbob_hansen2009_noiseless}, chosen to cover a range of landscapes (unimodal, multimodal, separable, ill-conditioned, etc.).
Each function is $5$-dimensional and uses a search domain of $[-5,5]^5$. For each function, we designate $5$ instances as training (instances $1$ to $5$) and $10$ as testing (instances $6$ to $15$) which are unseen during evolution, following a \textit{train-test} scheme to evaluate generalisation of the evolved algorithms. The LLaMEA evolution is given a budget of $100$ algorithms (evaluations) per run, and we repeat each configuration for $5$ independent runs (with different random seeds (seeds 1--5)).

\textbf{LLaMEA Configurations}: We consider six distinct configurations of LLaMEA, denoted LLaMEA-1 through
LLaMEA-6. These configurations differ in their mutation prompt strategy and evolutionary parameters, as summarized in Table \ref{tab:llamea_variants}.

\begin{table}[!t]
\centering
\caption{Overview of the six LLaMEA configurations evaluated in this study.
Population size $\mu$, offspring size $\lambda$, and elitism refer to the evolutionary
parameters, while the right-most column sketches the search paradigm.}
\label{tab:llamea_variants}
\begin{tabularx}{\linewidth}{l l l c X}
\toprule
\textbf{Variant} & \textbf{Mutation prompt(s)} & $\mu/\lambda$ & \textbf{Elitism} & \textbf{Search paradigm}\\
\midrule
LLaMEA-1 & \emph{Refine \& simplify} & 4 / 12 & \xmark & Population, incremental simplification\\
LLaMEA-2 & \emph{Generate new algorithm} & 4 / 12 & \xmark & Population, pure exploration\\
LLaMEA-3 & Both above prompts & 4 / 12 & \xmark & Population, mixed explore–exploit\\
LLaMEA-4 & Both above prompts & 1 / 1  & \cmark & Elitist $(1+1)$ hill-climber\\
LLaMEA-5 & Adaptive-mutation prompt & 4 / 12 & \xmark & Population, self-adaptive mutation\\
LLaMEA-6 & Adaptive-mutation prompt & 1 / 1  & \cmark & Elitist $(1+1)$ with self-adaptation\\
\bottomrule
\end{tabularx}
\end{table}

The specific mutations prompts are:
\begin{description}

    \item[Refine and Simplify] \textit{"Refine and simplify the
     selected algorithm to improve it."}
     \item[Random-new] \textit{"Generate a new algorithm
 that is different from those tried before."}
    \item[Adaptive-mutation] Here the LLM has to follow a refine prompt with a dynamically specified percentage (sampled from a long tail distribution) on how much to mutate the code~\cite{yin2025controlling}.
\end{description}

All methods start from scratch with the LLM generating initial algorithms (e.g., a random algorithm from the prompt describing the optimisation task). The fitness of an algorithm is evaluated as the mean
performance across the $5$ training instances of each BBOB function (aggregated across
functions). Specifically,
we use the BLADE experiment setup with training instances to evolve algorithms, then report performance on test instances for final evaluation. Key settings like function evaluation budget per algorithm (we use $2000d$
evaluations as typical in BBOB, so $10\,000$ for $d=5$) and logging frequency are consistent across methods
for fairness.

\subsection{Behaviour Metrics}

To analyse \emph{how} each generated algorithm performs its search (not just how well it performs), we record the complete optimisation trace of each algorithm on each function instance. A trace is the sequence of objective values
(fitness evaluations) and search-space locations over time as the algorithm runs. From these traces, we compute a set of scalar behavioural metrics that capture different
aspects of the search dynamics. Our metric set is inspired by previous works on search behaviour analysis, but here adapted to continuous optimisation traces. 

For the metrics below, let us define the search domain and optimisation trace as follows:
Let \(\mathcal D = \prod_{k=1}^{d}[l_k,u_k] \subset \mathbb R^{d}\) be the search
        domain with lower bounds $l_k$ and upper bounds $u_k$ and \(X=\{x_1,\dots,x_N\}\subset\mathcal D\) the set of evaluated points (trace).

We group the metrics into four categories: 
\medskip

\noindent \textbf{Exploration \& Diversity} Measure how broadly the algorithm searches the space.
    \begin{itemize}
        \item \textit{Average Nearest-Neighbors Distance} (\textbf{NN-dist}): Mean distance between each evaluated solution and its nearest
neighbour in the same trace, indicating the spread of search points. Higher values mean the
algorithm tends to sample points far apart (diverse exploration). Proposed by \cite{nezami2024building} as Mean Distance from Prior Evaluations (MDPE) to measure the novelty of newly sampled points versus the already evaluated ones, here we use it only to measure the average distance for the full trace.
        \item \textit{Coverage Dispersion} (\textbf{Disp}): The dispersion of points relative to the search domain. We approximate the
volume of space covered by the visited solutions. 

        \newcommand{\disp}{\operatorname{disp}}
        The {\em coverage--dispersion} metric is defined as
        \[
          \disp(X) \;=\;
          \sup_{y\in\mathcal D}\;
          \min_{x\in X}\bigl\|\,y-x\,\bigr\|_2 .
        \]
        It equals the radius of the largest empty hypersphere that can be placed inside
        \(\mathcal D\) without containing any sample from~\(X\); smaller values imply
        better space-filling coverage.
 
        \item \textit{Average Exploration Percentage} (\textbf{Expl \%}):
        The fraction of iterations classified as exploratory moves. The metric was proposed in \cite{subburaj2024population}. We have adapted the metric to calculate it per batch instead of per iteration, to significantly speed up the calculation without loosing too much information. 

        For any finite set $S\subset\mathbb R^d$ define
\[
  D(S)\;=\;\frac{2}{|S|(|S|-1)}
            \sum_{\substack{p<q\\ x_p,x_q\in S}}
            \lVert x_p-x_q\rVert_2 ,
\]
the average pairwise Euclidean distance.
       With widths $w_k = u_k-l_k$ and mean width
$\bar w = \tfrac1d\sum_{k=1}^d w_k$,
the expected pairwise distance of two \emph{i.i.d.} uniform
samples in $\mathcal D$ is approximated by  
\[
  D_{\mathrm{rand}}
  \;=\;
  \bar w\,\sqrt{\frac{d}{6}} ,
\]
(cf.\ the closed-form expression for the unit hyper-cube).
        
        Partition the trace $X$ into $C=\lceil n/K\rceil$ consecutive chunks
        $S_0,\dots,S_{C-1}$ of size $K$.  With the diversity measure
        $D(\cdot)$ introduced earlier, the exploration score of chunk $c$ is
                
        \[
          E_c
          \;=\;
          \min\!\bigl\{\,100\,
                   \tfrac{D(S_c)}{D_{\mathrm{rand}}},\;100\bigr\}
          \quad\bigl(0\le E_c\le100\bigr).
        \]
        
        The \textbf{average exploration percentage} is
        
        \[
          \overline{E}
          \;=\;
          \frac1C
          \sum_{c=0}^{C-1} E_c .
        \]

        and the complementary \textbf{exploitation} percentage is
        \(100-\overline{E}\).

    \end{itemize}

\medskip

\noindent \textbf{Exploitation \& Intensification:} Measure the focus on local search around the best found solutions.
    \begin{itemize}
        \item \textit{Average Distance to Best} (\textbf{Dist$\rightarrow$best}): The mean distance of each evaluated solution to the best solution found so far \cite{vcrepinvsek2013exploration}. A lower average distance indicates the algorithm spends more time exploiting near the current best (intensification).
        \item \textit{Intensification Ratio} (\textbf{Inten-ratio}): The proportion of iterations where a solution is within a small radius of the best-so-far solution . We set the radius to 10\% of the search range by default. A higher intensification ratio means the search frequently samples near the known best (intensive exploitation).
        \item \textit{Average Exploitation Percentage} (\textbf{Eplt \%}): Complementary to \textit{exploration percentage} (see above).
    \end{itemize} 
\medskip

\noindent \textbf{Convergence Progress}: Metrics indicating the speed and magnitude of improvement over time.

\begin{itemize}
    \item \textit{Average Convergence Rate} (\textbf{Conv-rate}): The geometric mean of successive error reductions, following
    definition from He et al. \cite{7122298}. Average convergence rate < 1 indicates convergent behaviour (fitness error decreases on
    average each step); smaller average convergence rate  means faster convergence.
    \item \textit{Average Improvement} (\textbf{$\Delta$ fitness}): The mean improvement in (normalized) objective value on iterations that yielded an improvement. This captures how large the typical improvement step is when progress is
    made.
    \item \textit{Success Rate} (\textbf{Success \%}): The proportion of iterations that resulted in any improvement over the current best. A higher success rate means the algorithm frequently finds better solutions (which could indicate either an easy problem or effective search steps).
\end{itemize}

\medskip

\noindent \textbf{Stagnation \& Reliability}: Metrics diagnosing lack of improvement and search stability.
\begin{itemize}
    \item \textit{Longest No-Improvement Streak} (\textbf{No-imp streak}): The longest sequence of iterations with no improvement. This reveals whether the algorithm can get stuck for long periods.
    \item \textit{Last Improvement Fraction} (\textbf{Last-imp frac}): The fraction of the total evaluations that have elapsed since the last improvement. A value close to 1 means the algorithm's final phase was stagnant, whereas a
    lower value indicates it kept improving until near the end.
\end{itemize}

These metrics are computed directly from each
algorithm’s trace data. They are fast to compute and 
together they span a multi-dimensional \emph{behaviour space}
that lets us position any LLM-generated meta heuristic run as a single
point~\footnote{%
We purposely restrict the study to single–objective, box-constrained
problems where the entire evaluation trace is available.}. For multi-instance evaluation, we aggregate the metrics computed per instance. Implementations of the metrics are provided in our repository  \cite{anonymous_2025_15675581}.

\subsection{Visualisation and Analysis Techniques}

To interpret the high-dimensional behavioural data and the evolution of the algorithms, we apply several
complementary visualisation techniques:

\textbf{Performance Convergence Plots}: For each LLaMEA variant, we evaluate the AOCC \cite{hansen2022anytime} anytime performance of all the generated algorithms. We aggregate
the results of the $5$ runs per LLaMEA variant by computing the mean best AOCC at each evaluation step. A higher AOCC means the generated algorithm finds good optima faster on average. We plot the mean best-so-far curves with confidence intervals from the $5$ runs. These curves allow visualising not just the final outcome but the trajectory of progress: e.g., whether a method quickly gains
moderate performance then plateaus, or improves steadily.

\textbf{Parallel Coordinates Plot for Metrics}: One way to view behaviour metrics is via a parallel coordinates plot, where each metric is an axis and each algorithm is represented by a poly-line across axes. We use
this to compare the metric profiles of different runs, especially to contrast high-performing versus low-performing cases. By coloring runs according to their final performance quartile, we identify which
metrics correlate with success. For instance, we will see if the top-performing runs tend to have consistently higher exploitation or lower stagnation than poorer runs.

\textbf{Code Evolution Graphs}: To visualise how each algorithm's code evolved during LLaMEA runs, we construct CEGs. In a CEG, each node represents a generated algorithm (with its code features and performance), and directed edges link a parent algorithm to its mutated
offspring. We extract a rich set of static code features from each algorithm’s source code via AST analysis, including counts of AST nodes and edges, graph connectivity measures (degree distribution, clustering coefficients), code complexity metrics (cyclomatic complexity, code token count, parameter counts), etc. We plot the total token count of the algorithm's code over generations as a curve to indicate code growth or simplification. The CEG figure shows how algorithms structurally evolve as LLaMEA progresses. Node sizes
in the CEG can be scaled by parent frequency; we size each node by how
many times it was selected as a parent, to highlight influential ancestors. We generate CEGs for each
LLaMEA variant for each independent run, arranging them in a grid for visual comparison.

%\go{Added a section about STNs below}

\textbf{Search Trajectory Networks}: To visualise the search dynamics of the 6 LLaMEA variants, we constructed STNs aggregating nodes and edges from the 5 runs of each variant. Nodes represent locations in the behaviour space, which we defined as a vectors of 5 real numbers representing the least correlated behaviour metrics: \textit{Expl. \%}, \textit{Conv-rate}, \textit{$\Delta$ fitness}, \textit{Success \%}, \textit{No-imp streak}. The behaviour space is partitioned into hypercubes  of a prefixed dimension. We explored two hypercube dimensions $\{0.01, 0.1\}$.  All sampled behaviour vectors within the same hypercube are grouped into a single node. Counts are kept for nodes and edges indicating their sampling frequency during the construction process. 

By combining these analyses, metric projections, code evolution visualisation, performance curves, and
trajectory networks, we obtain a multifaceted understanding of each method's behaviour.

\section{Results}

%\ns{Add the plots}

\subsection{Any-time Performance Comparison}
\begin{figure}[!b]
    \includegraphics[width=0.57\linewidth, valign=t,trim=0mm 0mm 0mm 0mm,clip]{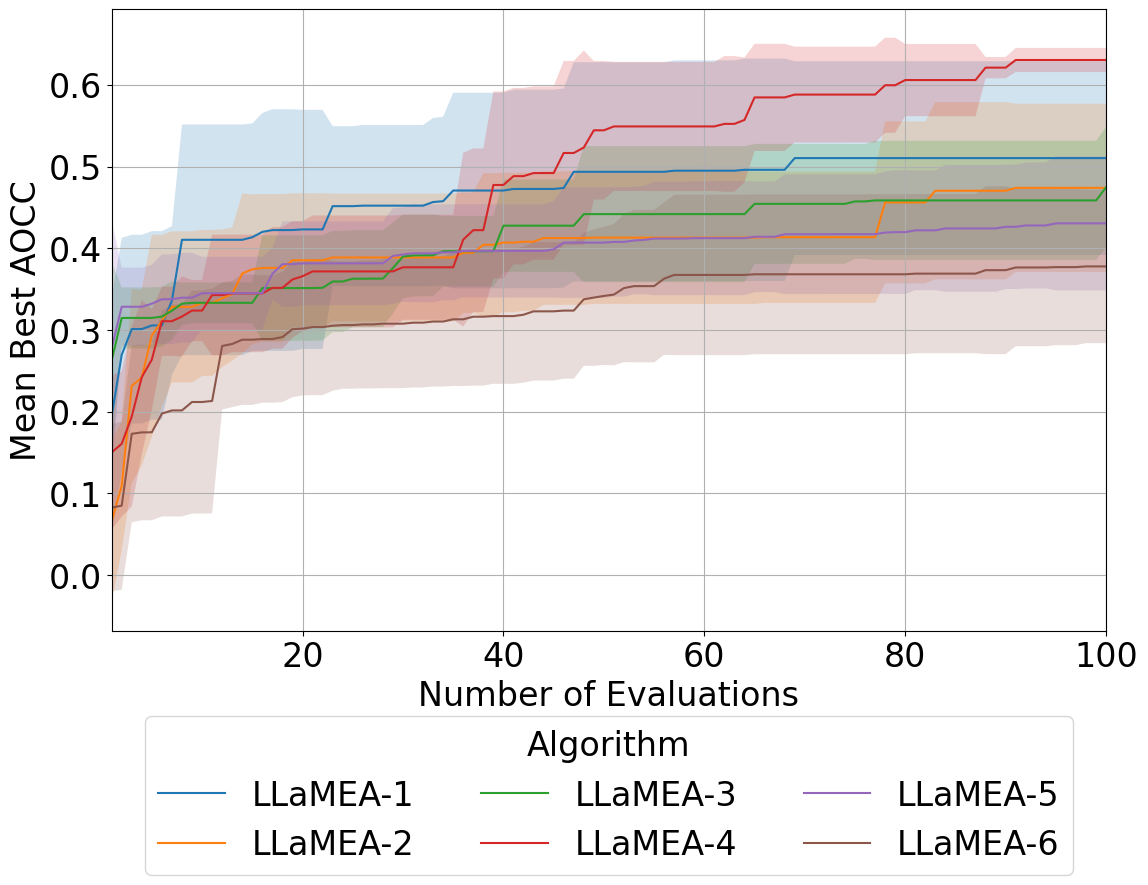}
    \includegraphics[width=0.41\linewidth, valign=t,trim=0mm 0mm 0mm 0mm,clip]{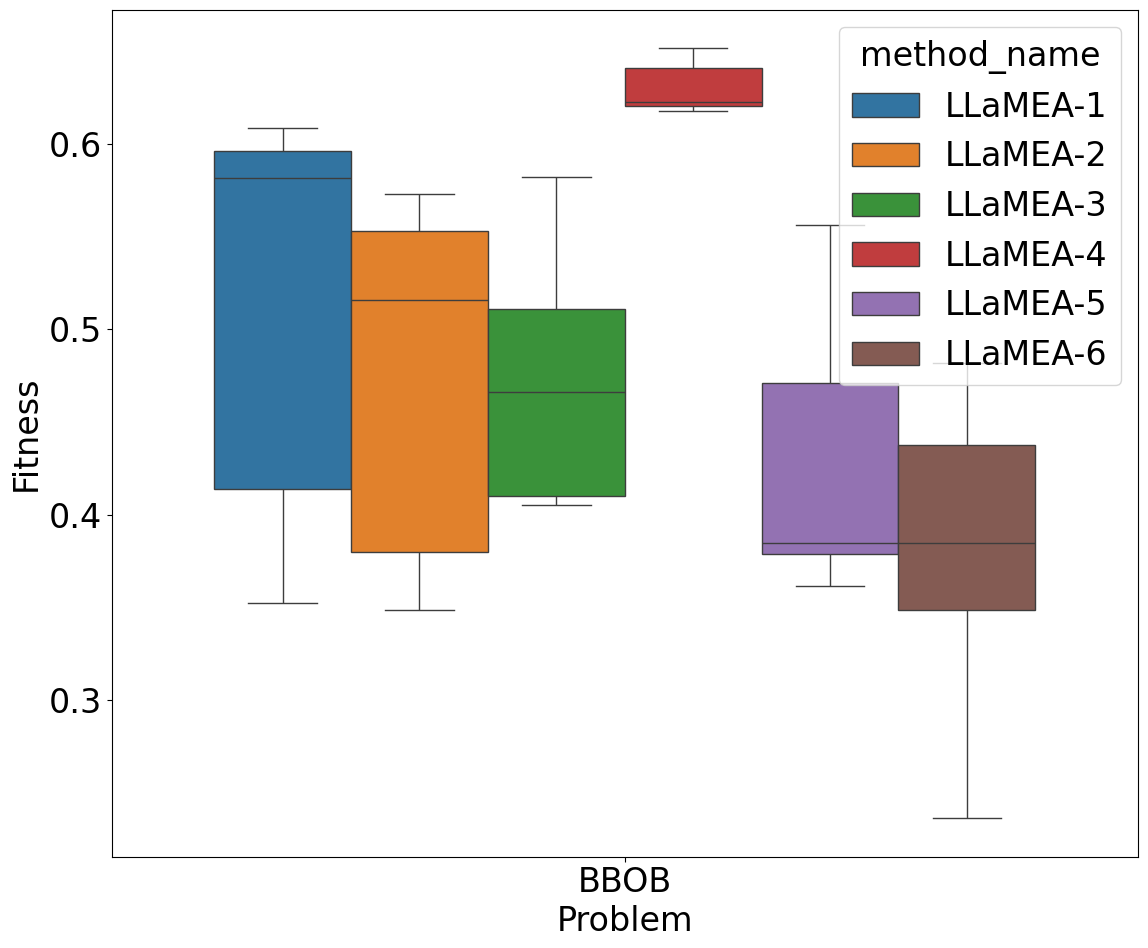}
    \caption{Aggregated results of the $5$ runs per LLaMEA variant by computing the mean best AOCC at each evaluation step (left subplot). A higher AOCC means the generated algorithm finds good optima faster on average. 95\% confidence intervals are visualized as shaded area. On the right the AOCC distribution of the $5$ final generated algorithms per LLaMEA configuration, evaluated on the $10$ test instances. \label{fig:aocc}}
\end{figure}

In Figure \ref{fig:aocc}, we can observe the performance of each LLaMEA variant aggregated over $5$ independent runs. It is interesting to observe that most variants show a relatively large variation between runs except for LLaMEA-4, which outperforms the other configurations. LLaMEA-4 uses both the `simplify' and `new random' mutation prompts using a $(1+1)$ strategy. Also from the validation results on the $10$ test instances (\ref{fig:aocc} right plot), it seems that the LLaMEA-4 variant was more stable and produced overall better algorithms. 

\subsection{Code Evolution and Search Dynamics}

\begin{figure}[!t]
    \includegraphics[width=0.96\linewidth,trim=0mm 0mm 0mm 0mm,clip]{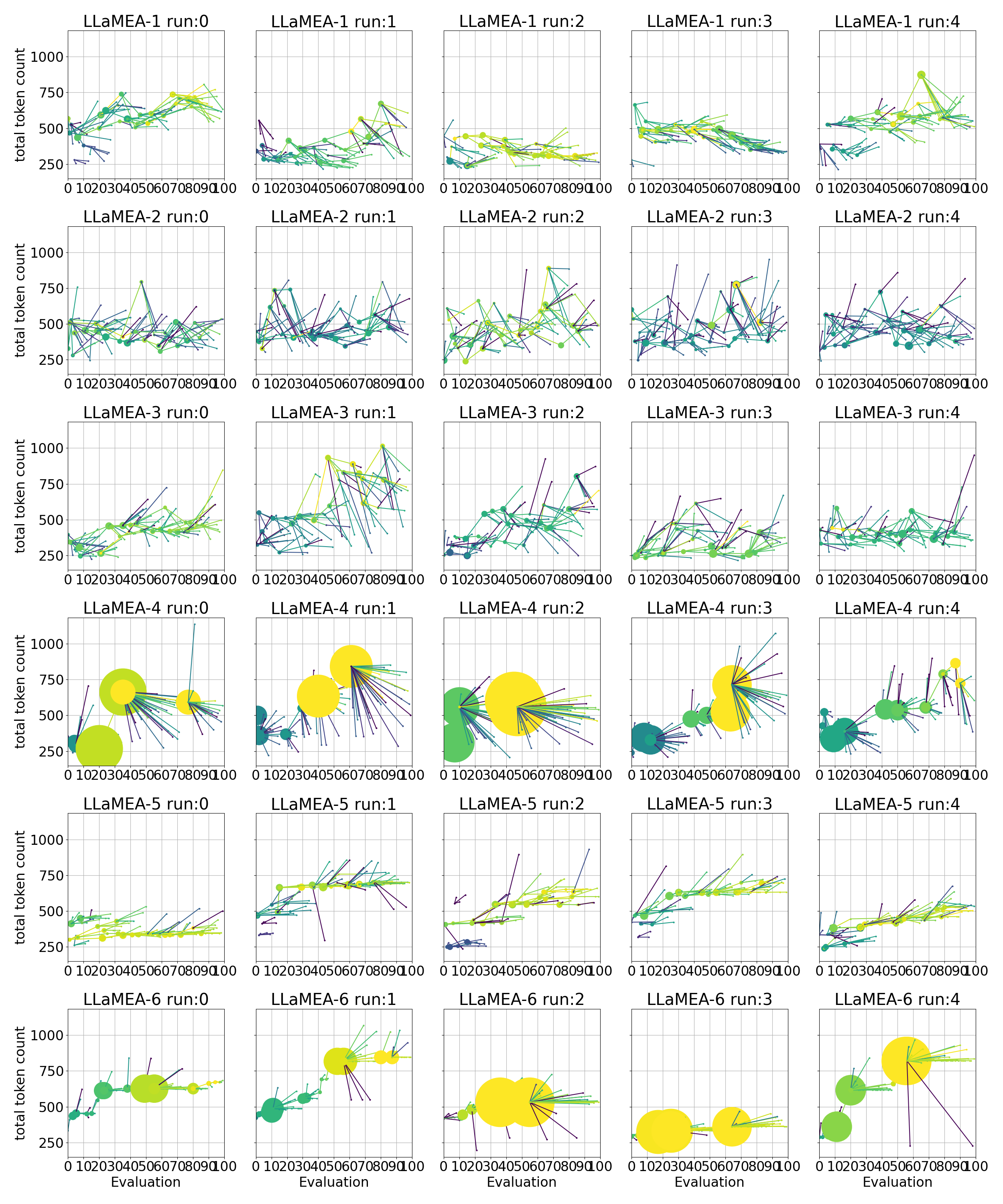}
    \caption{CEGs for the BBOB runs under different LLaMEA configurations (LLaMEA 1–6). Each sub panel corresponds to one independent run (columns) of a given configuration (rows). Within each panel, the y-axis shows the algorithm’s code length (total code token count) over the sequence of generated algorithms (x-axis), and the nodes denote the generated algorithms. Directed edges connect each algorithm to its offspring. Elitist 1+1 methods (LLaMEA-4 and 6) produce linear chains, while population-based methods yield branching graphs. Node colour denotes normalized AOCC (performance),  with bright yellow as the best AOCC and dark blue as the worst. \label{fig:ceg}}
\end{figure}

From Figure \ref{fig:ceg}, we can observe that LLaMEA-1 runs show relatively stable or even decreasing token count as the LLM simplifies code. LLaMEA-2 (new random) shows  high variance in code complexity between successive algorithms, indicating inconsistency in the generated algorithms. LLaMEA-5 and 6 (adaptive) show relative smaller updated in behaviour, sometimes simplifying, sometimes complexifying, but with small steps. Overall, CEGs reveal that the best-performing configuration (LLaMEA-4) managed to avoid code bloat while consistently improving algorithm performance (yellow colour), whereas others often grew in complexity without proportional gains.

\subsection{Behaviour Analysis}

%\ns{Add correlation matrix with numbers and explain which features we removed.}

\begin{figure}[!b]
    \includegraphics[width=0.8\linewidth,trim=0mm 0mm 0mm 0mm,clip]{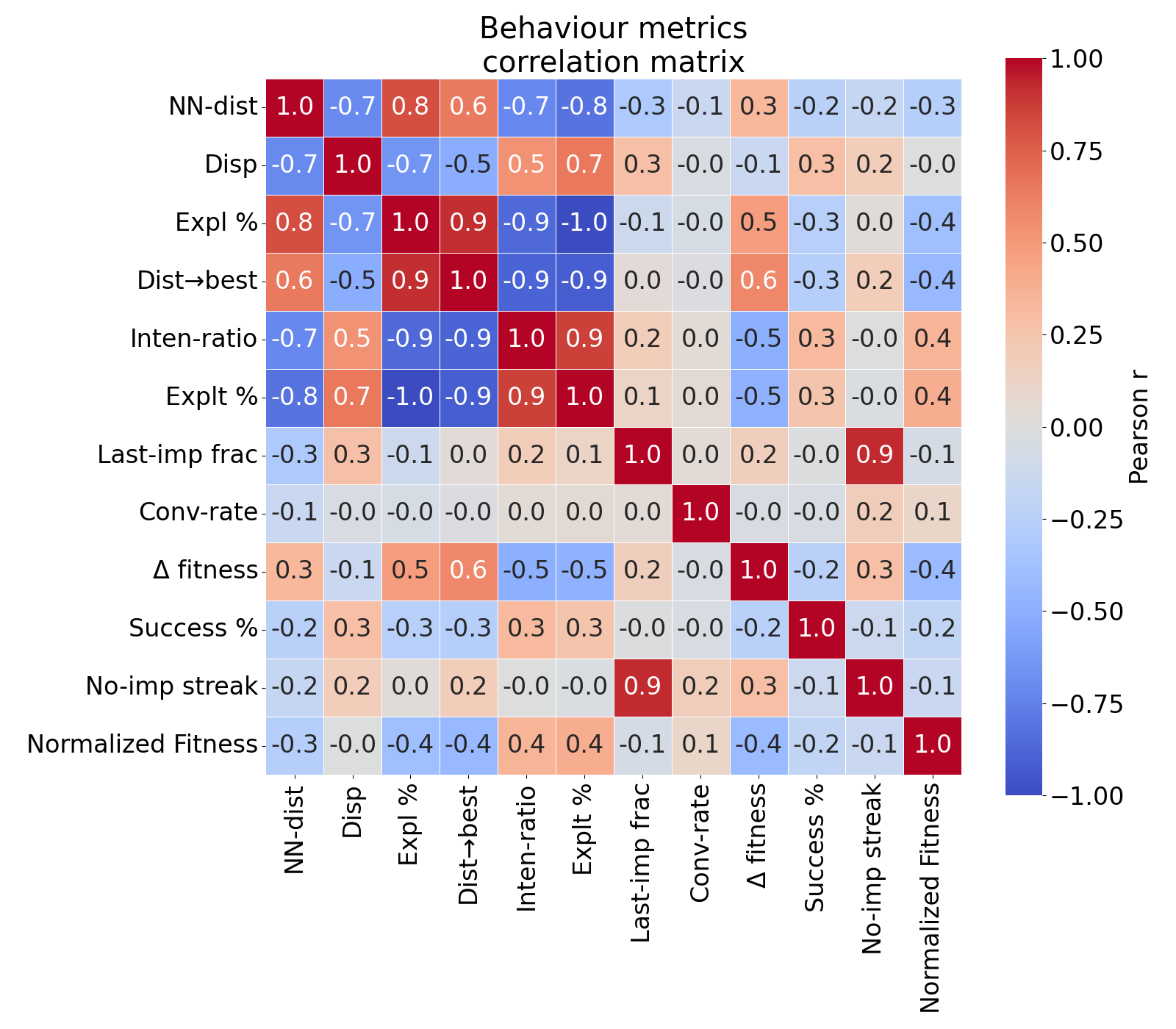}
    \caption{Pearson $r$ correlation between the different behaviour metrics and normalized fitness.\label{fig:corr}}
\end{figure}

\begin{figure}[!t]
    \includegraphics[width=0.49\linewidth,trim=0mm 0mm 0mm 0mm,clip]{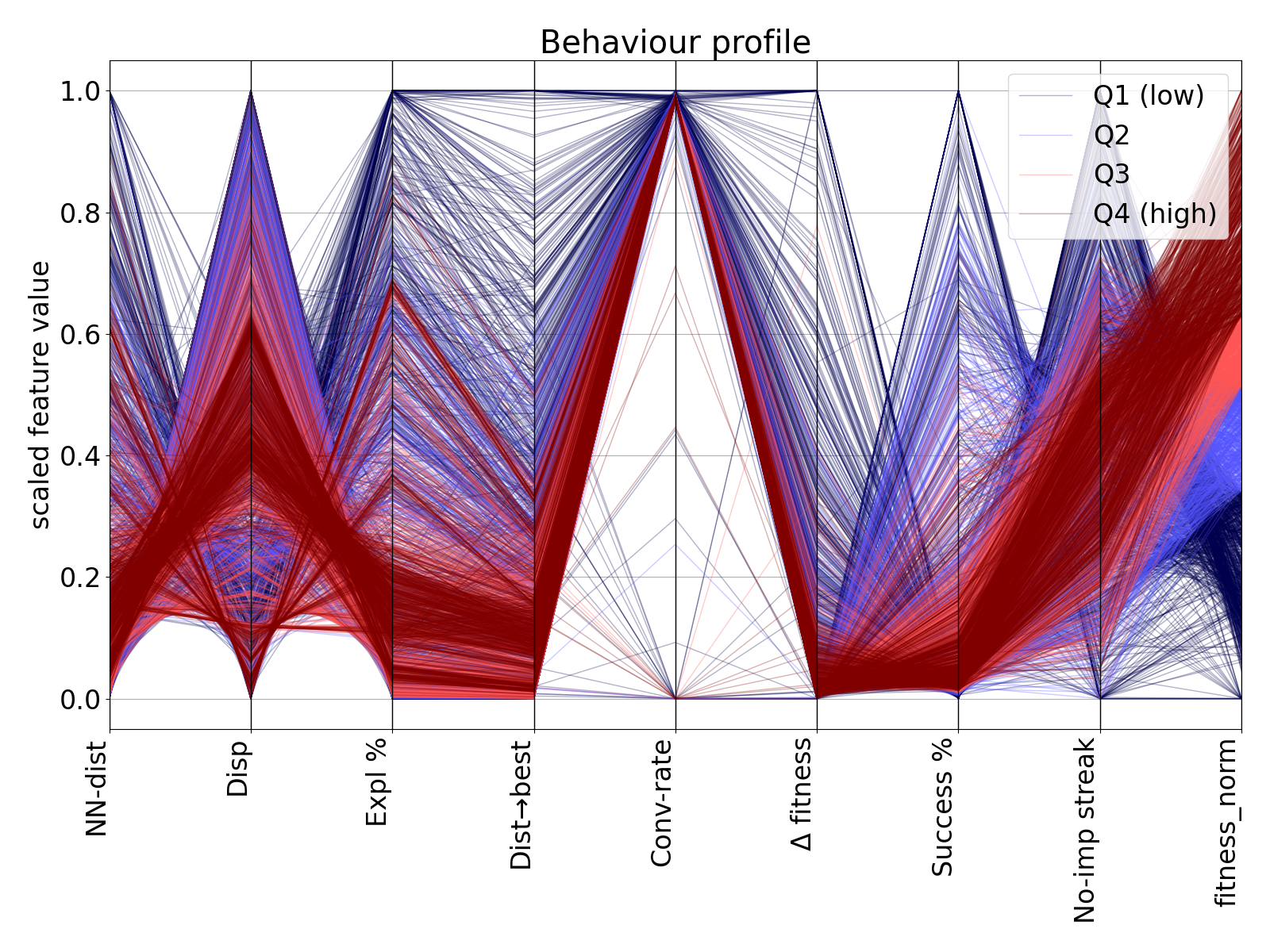}
    \includegraphics[width=0.49\linewidth,trim=0mm 0mm 0mm 0mm,clip]{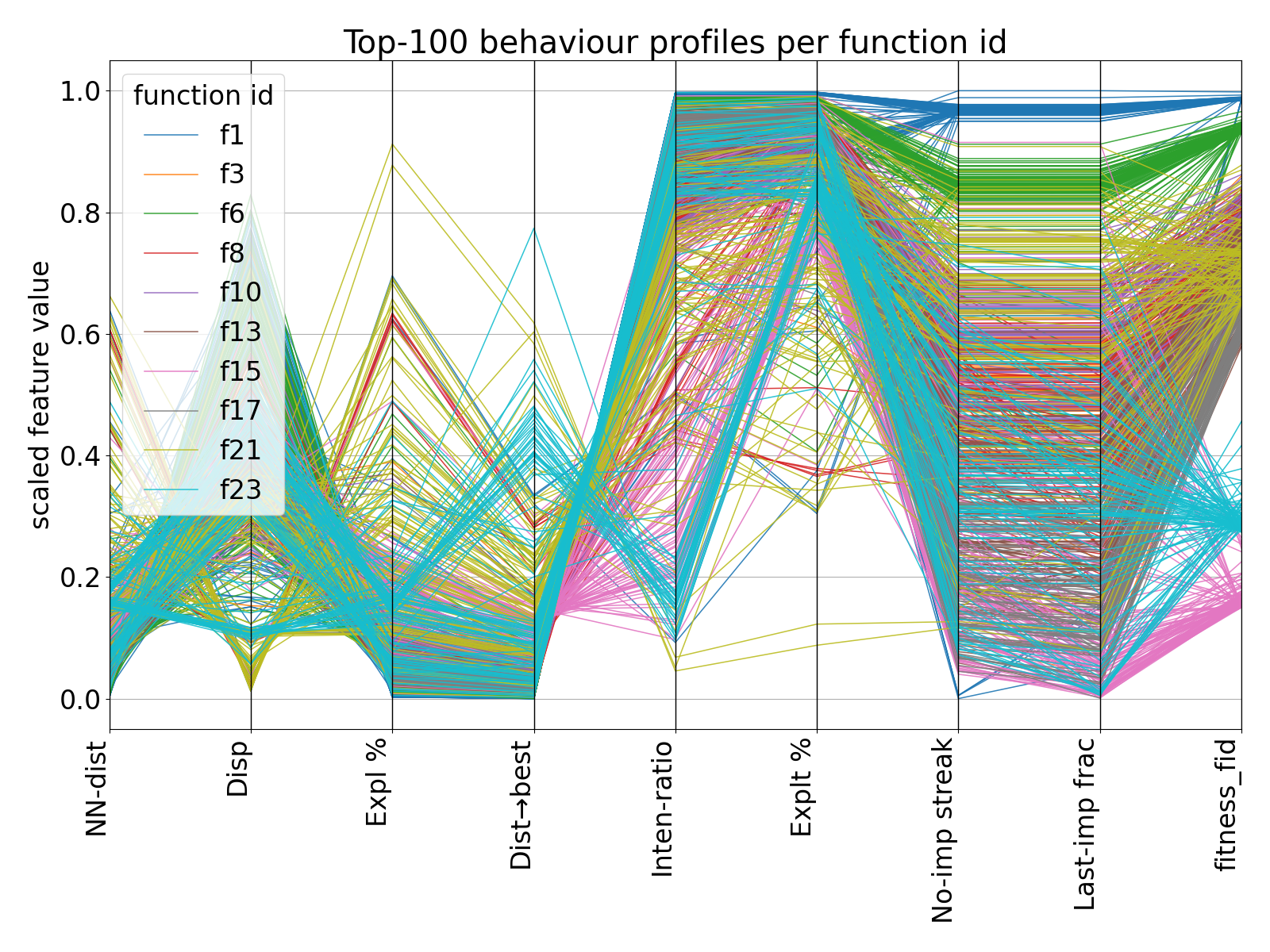}
    \caption{Behaviour profile over all generated algorithms (left) and the best $100$ algorithms per BBOB function (right).\label{fig:behaviour}}
\end{figure}

To verify that the defined behaviour metrics are complementary to our analysis, we started with a correlation analysis. In Figure \ref{fig:corr}, the Pearson $r$ correlation is given between all considered metrics and including the normalized fitness (AOCC) score. It can be observed that some features clearly correlate. The most obvious pair is \textit{Exploitation \%} and \textit{Exploration \%,} this was expected as it is by design. Other pairs include \textit{Expl.\%} with \textit{Distance to best}, \textit{Exploitation \%} with \textit{Intensification ratio} and \textit{Last no-improvement fraction} with \textit{No improvement streak}. For further analysis we have selected only one of these highly correlated feature pairs.
In Figure \ref{fig:behaviour} parallel coordinate plots of the normalized behaviour metrics for each generated algorithm are provided. On the left side we can see a clear ``profile'' for good algorithms (dark red), with an average dispersion, low nearest neighbour distance and relatively low exploration percentage. On the right side of the figure we can see the top performing algorithms per individual BBOB function. From this figure it is clear that some objective functions benefit from a different behaviour profile than others, while for some metrics such as exploration percentage this is relatively condensed, meaning that similar behaviour works well on all problems.

\begin{figure}[!tb]
    \includegraphics[width=\linewidth,trim=0mm 0mm 0mm 0mm,clip]{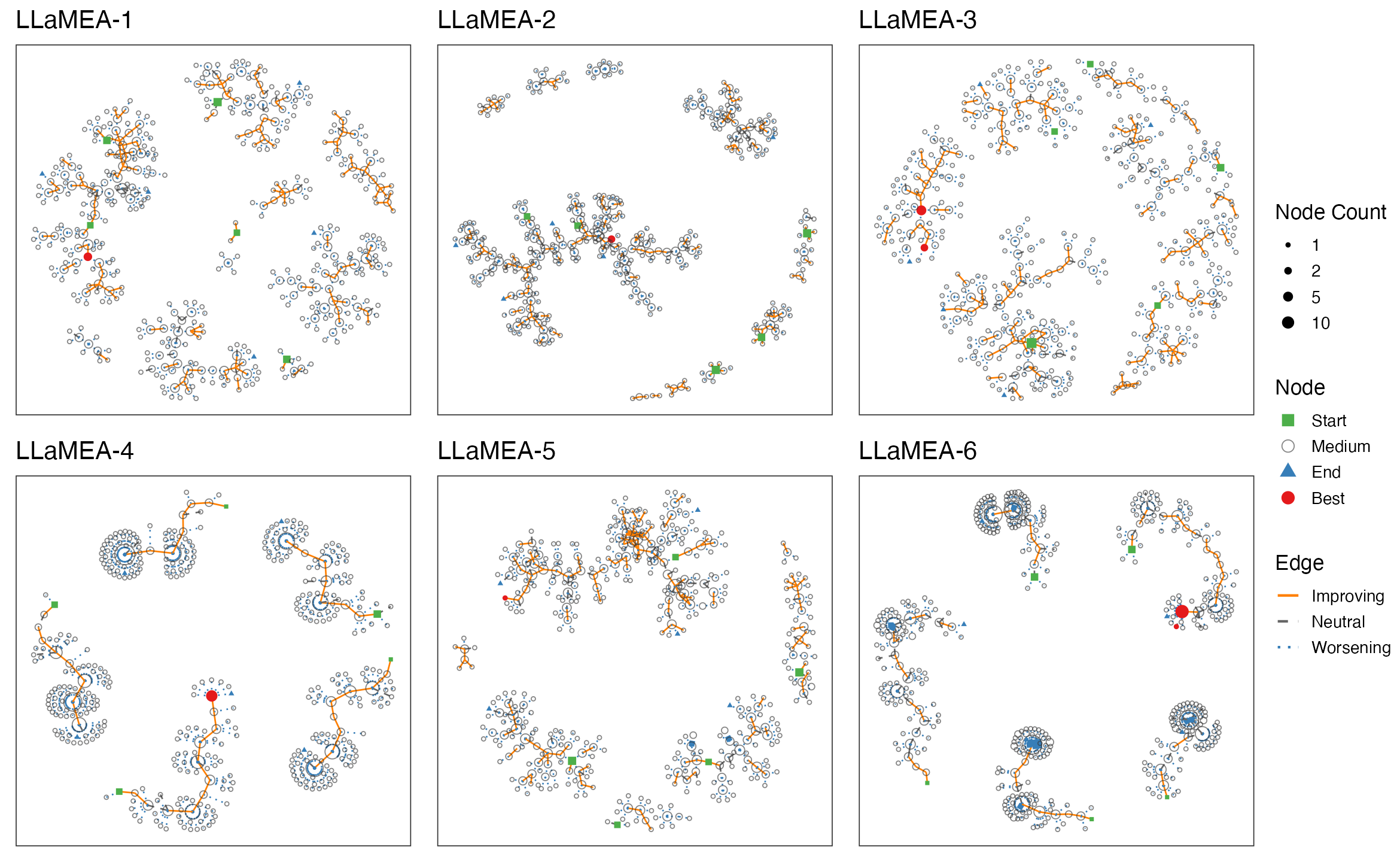}
    \caption{Search trajectory networks for the 6 LLaMEA variants with partition hypercubes of size 0.01. Node and edge types are as indicated in the legend. Node sizes and edge widths are proportional to sampling frequency. \label{fig:stn_001}}
\end{figure}

Figure \ref{fig:stn_001} shows the STNs for the 6 LLaMEA variants with a partition hypercube of dimension 0.01. For the STN analysis we have selected the $5$ least correlated metrics: \textit{Expl. \%}, \textit{Conv-rate}, \textit{$\Delta$ fitness}, \textit{Success \%}, \textit{No-imp streak}.
At this level of granularity, there is little convergence across runs and the networks feature several disconnected components. Nodes indicating the start of runs (green squares) appear in separated components for all variants except the first two, where two start nodes can be seen in the same connected component. This visualisation captures the different EAs selection dynamics. The population variants (LLaMEA-1, 2, 3 and 5) show ``bushy" graphs with wider branches of improving edges, as well as small components indicating solutions in the population that were not further explored in the evolutionary process.  On the other and, the 1+1 variants (LLaMEA 4 and 6) show elongated components (one for each run) following a single path of improving edges with side clusters of deteriorating edges indicating unsuccessful mutation attempts.

\begin{figure}[!tb]
    \includegraphics[width=\linewidth,trim=0mm 0mm 0mm 0mm,clip]{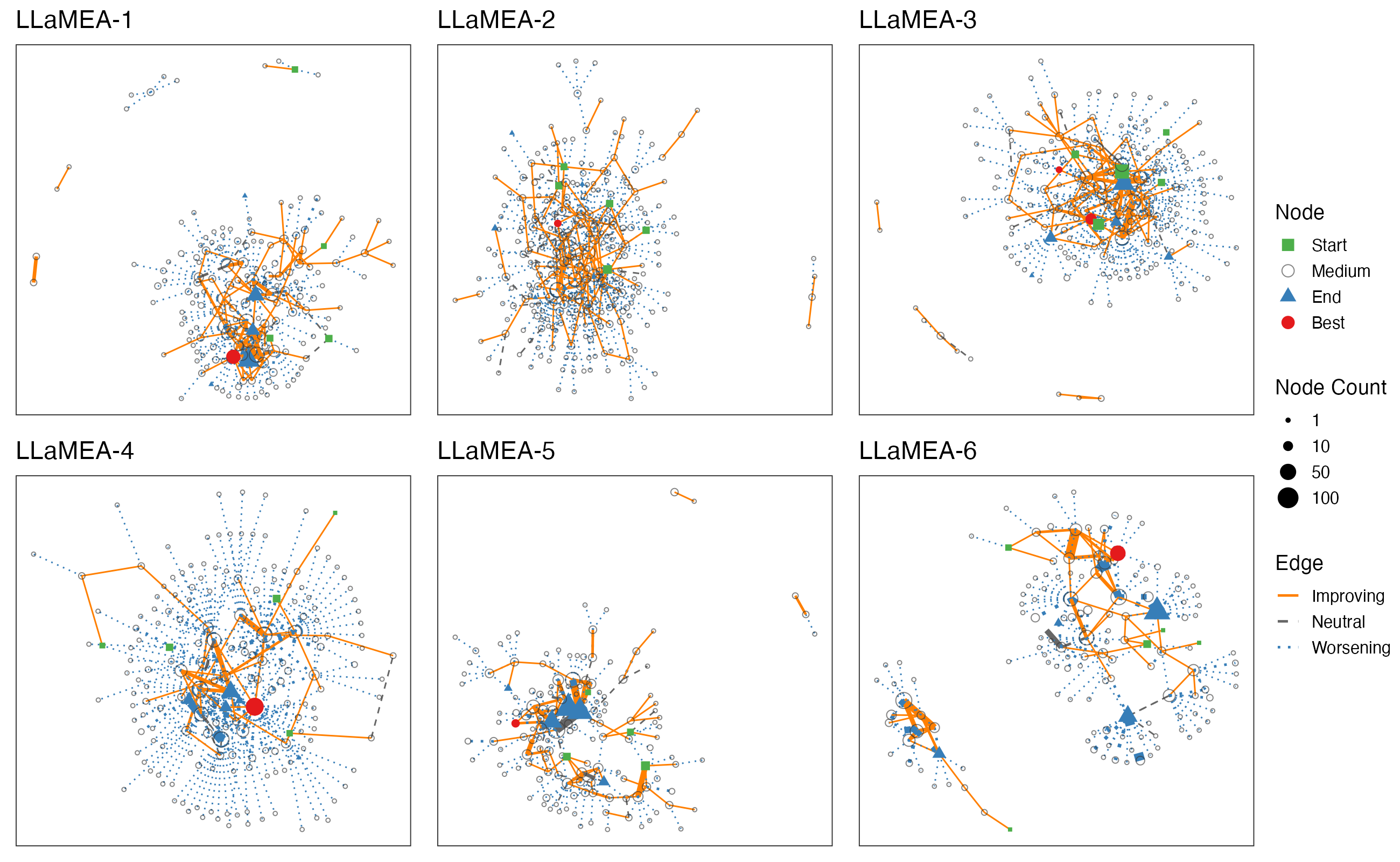}
    \caption{Search trajectory networks for the 6 LLaMEA variants with partition hypercubes of size 0.1. Node and edge types are as indicated in the legend. Node sizes and edge widths are proportional to sampling frequency. \label{fig:stn_01}}
\end{figure}

With the coarser partition factor, that is, larger hypercubes of dimension 0.1 for grouping solutions into nodes (Figure \ref{fig:stn_01}) we observe convergence across runs for most variants: a single large component is now visible. The exception is LLaMEA-6 where a second component of smaller size can be seen. The most explorative variant is LLaMEA-2 featuring the largest network (larger number of nodes and edges), while the most compact and efficient search is captured by the elitist variant LLaMEA-4, featuring short paths to the best solution.

%\go{Discuused STN plots, not sure how to update the Discussion.}

\section{Discussion}

Our findings highlight several important points for the design of LLM-driven algorithm generators and for the analysis of their behaviour: 

(1) \textit{The importance of exploitation-exploration balance in AAD}: Through LLaMEA-4’s performance, we see that giving the LLM two contrasting mutation prompts, one to exploit (refine/simplify the current best algorithm) and one to explore (generate something entirely new), yields a highly effective search. This mirrors classic evolutionary computation principles: too much exploration (LLaMEA-2) is inefficient, while too little (LLaMEA-1) risks premature convergence. 
The LLM’s creative capabilities might tempt one to always generate fresh algorithms, but our results demonstrate the value of refining what you have. Notably, the simplify prompt not only improved performance but often reduced code complexity, suggesting that simpler algorithms often generalized better and were easier for the LLM to tune. 

(2) \textit{The role of Elitism and selection pressure}: The stark difference between LLaMEA-3 (both prompts, no elitism) and LLaMEA-4 (both prompts, elitist 1+1) indicates that how we select and carry forward algorithms in each generation has a significant effect. LLaMEA-4 effectively conducts a $(1+1)$-ES in the space of algorithms, which guarantees never losing the best found. This appears important in the AAD context because evaluating each algorithm is expensive; one cannot afford to discard a good strategy hoping it might re-evolve later. The population-based methods occasionally lost their best solutions when offspring didn’t include similar quality algorithms. That said, maintaining diversity is still important; an interesting future direction is to incorporate niching or maintaining a portfolio of top algorithms.

(3) \textit{Interpretability through behaviour metrics}: By using explainable behaviour metrics, we were able to diagnose why certain methods underperformed, something that would be difficult from performance data alone. For instance, we could attribute LLaMEA-2’s poor performance to its low success rate and high stagnation, pointing to an overly exploratory behaviour. 
We also learned that in general, algorithms that obey to a certain behaviour profile, generally perform better on BBOB problems than others. This includes a relatively low Nearest Neighbour distance (small step size), a exploitation focused strategy and a relatively low success chance ($0.1$). We also could observe that for different BBOB functions, different behaviour patterns are beneficial, though many of the BBOB functions seem to share at least some portion of the behaviour profiles in their top-performers.

(4) \textit{Limitations}: It should be noted that the study is currently limited to relatively low-dimensional problems (5D) and one type of LLM; behaviour on higher dimensions or with different LLMs (e.g., coding styles of different models might differ. For example, a larger LLM might generate more complex code by default or conversely, handle instructions to simplify even better. The still limited number of $5$ runs also means that the conclusions of these work are made with caution. Due to the API and computational expenses we could not afford additional runs for now.

%\ns{Add the discussion draft}

\section{Conclusions and Future Directions}

We presented an in-depth behaviour space analysis of meta-heuristics generated by an LLM-driven evolutionary algorithm (LLaMEA) on black-box optimisation tasks. 
By defining a rich set of behavioural metrics and visual representations, we were able to explain performance differences between various LLaMEA configurations that use different prompt-based mutation strategies. Our analysis identified a particular configuration, using dual prompts for simplification and diversification in a 1+1 elitist framework (LLaMEA-4), as the most successful. By analysing the behaviour space of these algorithms we identified an effective exploration-exploitation balance. From a methodological perspective, this work demonstrates the value of combining trace analysis (exploration/exploitation metrics, STNs) with static code analysis (CEGs) to get a full picture of algorithm evolution. We showed that LLM-generated algorithms \textit{can} be analysed with similar tools developed for traditional meta-heuristics, and these tools can yield actionable insights. In practical terms, an engineer using LLaMEA or a similar system could use our approach to diagnose why an automated design run failed (e.g., metrics show it stagnated early) or succeeded (e.g., code complexity dropped, indicating a cleaner strategy was found), and adjust the system accordingly (change prompts, selection, etc.). 

There are several avenues for future work. First, applying this behaviour analysis to other problem domains (e.g., combinatorial problems or mixed-integer optimisation) would test the generality of the findings. The STN and attractor concepts, for example, could reveal different patterns if the search space has discrete states. Second, integrating these analysis techniques into the evolutionary loop is a promising direction, for instance, one could create a feedback where if the behaviour metrics indicate too much exploitation, the system automatically increases exploration pressure. This would effectively close the loop between analysis and design, leading to self-correcting LLM-driven optimizers. Third, scaling to other LLMs and more complex problems will be important. While the BBOB suite gave a meaningful test-bed, real-world optimisation often involves constraints, high dimensions, or multiple objectives. Understanding how an LLM designs algorithms under those conditions, and whether our identified best practices hold, is an open question. 

Finally, from an explainable AI standpoint, one could explore user-friendly visualisations (perhaps interactive) based on our approach, to allow human experts to inspect and verify automatically designed algorithms before trusting them on critical tasks. In conclusion, the synergy of LLMs with evolutionary search opens exciting possibilities for automatic algorithm discovery. By mapping the behaviour space of these evolved algorithms, we gain not only performance improvements but also understanding, a crucial step toward reliable and accountable AI-driven algorithm design. We hope this study serves as a template for analysing complex adaptive systems that include AI components, and that the insights gained will guide both theorists and practitioners in harnessing LLMs for optimisation in a principled way.

\subsubsection*{Disclosure of Interests.}
The authors have no competing interests to declare that are relevant to the content of this article.

\FloatBarrier

%\ns{Add the conclusions / future work}

\bibliographystyle{splncs04}
\bibliography{main}

\end{document}